\documentclass[runningheads]{llncs}
\usepackage{graphicx}
\usepackage{url}
\usepackage{subcaption}
\usepackage{arydshln} 

\begin{document}
\title{A Knowledge Graph for Assessing\\Aggressive Tax Planning Strategies}
\author{Niklas L{\"u}demann\inst{1} \and Ageda Shiba\inst{1} \and Nikolaos Thymianis\inst{1} 
\and\\Nicolas Heist\inst{1}\orcidID{0000-0002-4354-9138} \and Christopher Ludwig\inst{2,3}\orcidID{0000-0001-8268-2196} \and Heiko Paulheim\inst{1}\orcidID{0000-0003-4386-8195}}

\authorrunning{N. L{\"u}demann et al.}

\institute{Data and Web Science Group, University of Mannheim, Germany\\\email{\{nluedema,agshiba,nthymian\}@mail.uni-mannheim.de},\\\email{\{nico,heiko\}@informatik.uni-mannheim.de} \and
Area Accounting and Taxation, University of Mannheim, Germany
\and
Corporate Taxation and Public Finance Department\\Leibniz Centre for European Economic Research, Mannheim, Germany\\\email{christopher.ludwig@zew.de}}

\maketitle              

\begin{abstract}
The taxation of multi-national companies is a complex field, since it is influenced by the legislation of several states. 
Laws in different states may have unforeseen interaction effects, which can be exploited by allowing multinational companies to minimize taxes, a concept known as \emph{tax planning}.
In this paper, we present a knowledge graph of multinational companies and their relationships, comprising almost 1.5M business entities.
We show that commonly known tax planning strategies can be formulated as subgraph queries to that graph, which allows for identifying companies using certain strategies.
Moreover, we demonstrate that we can identify anomalies in the graph which hint at potential tax planning strategies, and we show how to enhance those analyses by incorporating information from Wikidata using federated queries.

\end{abstract}

\section{Introduction}
Multinational corporations (MNCs), such as Google, IKEA, and Apple, have been scrutinized in the recent decade for so-called ``aggressive'' tax planning strategies. Taxes have a considerable effect on the net income of corporations, and it is in principle in the best interest of MNCs to reduce their worldwide tax burden by relocating profits within their group to lower-taxed affiliates.

The increasing internationalization of business activities in combination with the growing importance of the digital economy can create conflicts for the taxation of business profits by local governments \cite{oecd_addressing_2015}. For cross-border businesses' activities, an appropriate allocation of foreign and domestic profits – and the underlying capital – to the involved jurisdictions is necessary, in accordance with the principle of economic allegiance \cite{jacobs_internationale_2016}. MNCs represent an economic entity, but they are usually organized as a conglomerate of legally independent separate legal entities or permanent establishments. The direct method to allocate profits and costs follows the separate entity approach and requires corporate divisions to behave as independent market participants, whereas the indirect method follows the unitary entity approach and allocates profits to affiliates by a formulary apportionment. The prevailing method in the international tax system is both for separate legal entities and permanent establishments the direct method which requires the application of the arm’s length principle to intra-group transactions \cite{oecd_model_2014}. However, for many intermediary goods, services, and license contracts within MNCs, no independent reference market is observable and the implementation of the arm’s length principle can be difficult. 

Intuitively, MNCs have an incentive to allocate profits and costs in a tax-efficient way to reduce the overall tax burden of the corporation \cite{endres_international_2015}. Tax reduction has a positive effect on the consolidated net income of MNCs which increases shareholder value. Efficient tax systems are – in theory – required to be neutral regarding any investment decision, but the diverse application of international taxation principles leads to a considerable heterogeneity between national tax systems \cite{devereux_general_1991}. Taxes represent costs for corporations, thus, MNCs usually consider tax effects intensively and pursue substantive and formal tax planning activities to change and structure economic activities in a tax-efficient way. 

The term \emph{tax planning} refers to generally accepted strategies to minimize tax liabilities of MNCs. Up to now, it is not precisely defined which tax planning strategies are considered as ``aggressive''. The Organization for Economic Co-operation and Development (OECD) defines them as planning activities with ``unintended and unexpected tax revenue consequences'' \cite{oecd_study_2008}. In general, ``aggressive'' tax planning strategies are said to be in line with legal provisions but these strategies might be able to considerably reduce the tax burden of MNCs in some regions. In the following, the term ``aggressive'' refers to legal tax planning strategies of MNCs that lead to a substantial reduction of their tax liabilities \cite{heckemeyer_masnahmen_2013}. Tax planning has to be differentiated from the terminology of \emph{tax avoidance} and \emph{tax evasion}. Tax avoidance strategies exploit loopholes in the tax law to reduce the tax liability. Tax evasion refers to any illegal activities to minimize the tax burden (e.g. misstatements in the tax declaration) \cite{endres_international_2015}.

MNCs are usually not one business entity, but a network of parent and child companies and holdings across different countries. Therefore, they can be directly represented in a knowledge graph (KG) \cite{ehrlinger2016towards}, i.e., a graph describing entities and their relations \cite{paulheim2017knowledge}. In such a KG, companies can be connected among each other as well as to the countries they belong to, and further information (such as companies' legal forms, countries' populations and GDP etc.) can be added. Such a KG allows for two kinds of analyses: First, companies using certain aggressive tax planning strategies can be identified in the graph, since they correspond to characteristic subgraph patterns. Second, the graph can be analyzed for anomalies, which might hint at tax avoidance strategies, which are not yet known.

The rest of this paper is structured as follows. Section~\ref{sec:kg} describes the knowledge graph used for our analysis and its sources. Section~\ref{sec:usage} demonstrates the above mentioned use cases, i.e., the identification of aggressive tax planning strategies and the search for graph anomalies. Section~\ref{sec:related} discusses relevant related work, and section~\ref{sec:conclusion} closes with a summary and an outlook on future work.

\section{Knowledge Graph}
\label{sec:kg}
For our analysis, we combine data from different sources into a knowledge graph, which can then be queried for analytics purposes.

\subsection{Data Sources}
The main source of our KG is the Global Legal Entity Identifier Foundation\footnote{\url{https://www.gleif.org/en/}}. GLEIF collects data from different legal entity identifier (LEI) issuers and provides a consolidated collection of that data. For each legal entity, different data fields (such as address, legal form, etc.) are collected. GLEIF has two levels of data: level 1 data (\emph{who is who}) contains data about the companies as such, whereas level 2 data (\emph{who owns whom}) provides information about the relationships between companies.

The level 2 data contains both direct as well as ultimate subsidiaries, i.e., child companies of child companies and so on. The latter is, in theory, equivalent to following the transitive closure of the subsidiary relation, however, in some cases, there are subsidiaries missing in between in the data for various reasons (e.g., country specific regulations for disclosing that information).

For further analyses, we include economic and geographic data for the entities at hand. To that end, country-specific data from the World Bank\footnote{\url{https://data.worldbank.org/}} and Wikidata~\cite{vrandevcic2014wikidata} is collected. Those country-wide indicators include population and GDP. Moreover, we included the statutory corporate tax rate for each country from the OECD corporate tax database\footnote{\url{https://www.oecd.org/tax/tax-policy/corporate-tax-statistics-database.htm}}.

Since some data was imported from Wikidata, we also provide interlinks to Wikidata. Countries and companies were trivial to match, since for the former, the GLEIF dataset uses ISO codes also present in Wikidata\footnote{\url{https://www.wikidata.org/wiki/Property:P297}}, whereas for the latter, GLEIF identifiers are also  used in Wikidata.\footnote{\url{https://www.wikidata.org/wiki/Property:P1278}}. Using that approach, we could interlink all countries and a total of 20,734 companies to Wikidata.

For matching cities, first, candidates are retrieved from Wikidata based on postal codes. To that end, a list of all entities with postal codes was retrieved from Wikidata, and attribute values with ranges are preprocessed to get an actual map of postal codes to entities (e.g., Berlin has only one value for the postal code attribute with value \verb+10115-14199+\footnote{\url{https://www.wikidata.org/wiki/Q64}}). To deal with entities that do not represent a city (e.g., streets or libraries) and with cases where multiple candidates exist (e.g., \texttt{1000} is the postal code for Brussels, Sofia, Ljubljana, among others), the matching was made based on edit distance, with a maximum threshold of 0.3. Using that approach, we were able to link 43,832 cities to Wikidata.\footnote{The full code for generating the knowledge graph is available online at \url{https://github.com/tax-graph/taxgraph}.}

One basic design decision is collecting the data in one knowledge graph, vs. using SPARQL federated queries for Wikidata and Worldbank data. After some initial experiments with Virtuoso's query federation functionality, we found that federated queries are possible, but significantly slower than local queries. Hence, we follow a mixed approach: data about central entities (such as the population and GDP for countries) are included in our knowledge graph, while still maintaining the possibility to use the full data in Wikidata via federation.

\subsection{Resulting Graph}
\begin{figure}[t]
    \centering
    \includegraphics[width=\textwidth]{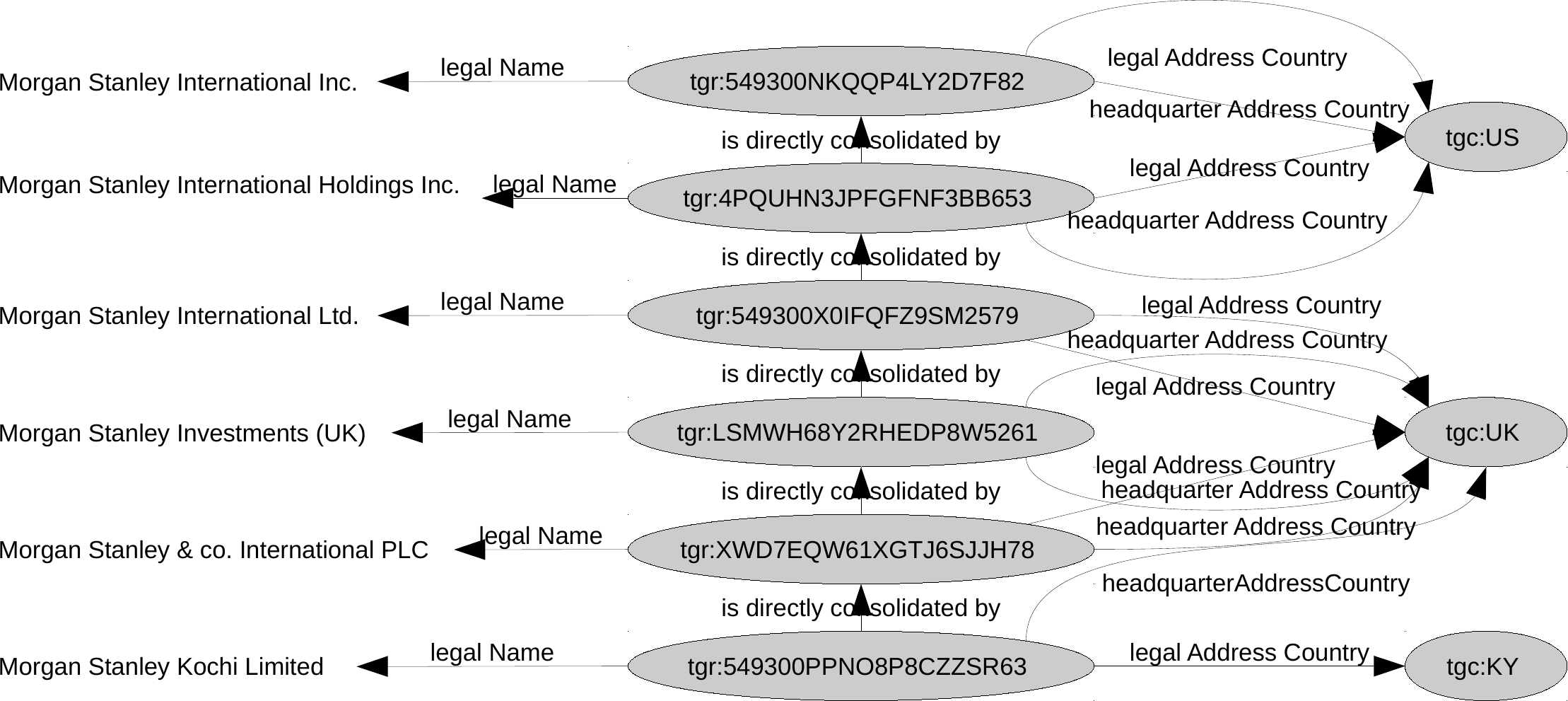}
    \caption{Example representation of a company and its direct parents in the taxation graph}
    \label{fig:example_sap}
\end{figure}

The resulting graph contains about 1.5M companies and 180k relationships between those companies, as shown in table~\ref{tab:kg_contents}. An example representation of a company is shown in Fig.~\ref{fig:example_sap}. Overall, the graph has 22,839,123 triples and is stored in a Virtuoso RDF store~\cite{erling2012virtuoso}. The knowledge graph is available online for browsing, download, and querying via a SPARQL endpoint.\footnote{\url{http://taxgraph.informatik.uni-mannheim.de/}}

\begin{table}[t]
    \caption{Contents of the Knowledge Graph}
    \label{tab:kg_contents}
    \centering
    \begin{tabular}{l|r}
         \emph{Class} &  \emph{Count} \\
         \hline
         Company & 1,491,143 \\
         Country & 225 \\
         City & 95,306 \\
         Legal Form & 1,286 \\
         \hline
         \emph{Relation} & \emph{Count} \\
         \hline
         direct subsidiary & 87,020 \\
         ultimate subsidiary & 96,465 \\
    \end{tabular}
\end{table}

\begin{figure}[t]
    \centering
    \includegraphics[width=\textwidth]{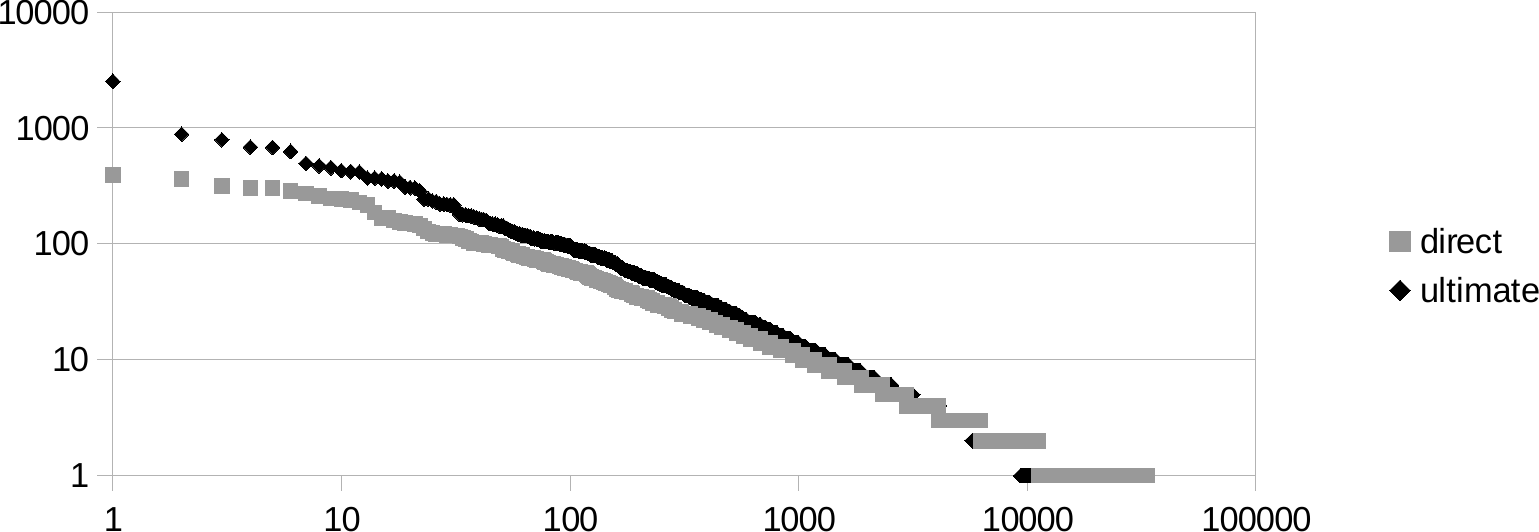}
    \caption{Distribution of the number of direct and ultimate children per company}
    \label{fig:distribution_ultimate_children}
\end{figure}

As depicted in Fig.~\ref{fig:distribution_ultimate_children}, the distribution of direct and ultimate children follows a power law distribution. There are a few companies with very high number of ultimate children, as shown in table~\ref{tab:top10_ultimate_children}, whereas the majority has only one or no ultimate children, as shown in Fig.~\ref{fig:comparison_kgs}. Companies with children have on average 2.6 direct children and 4.1 ultimate children (i.e., members of the transitive closure of the child relation). 
The longest chains of subsidiaries that we find spans across six companies, as shown in Fig.~\ref{fig:example_sap}: Here, the ultimate child has a legal address in the Cayman Islands.

\begin{figure}[t]
    \centering
    \includegraphics[width=\textwidth]{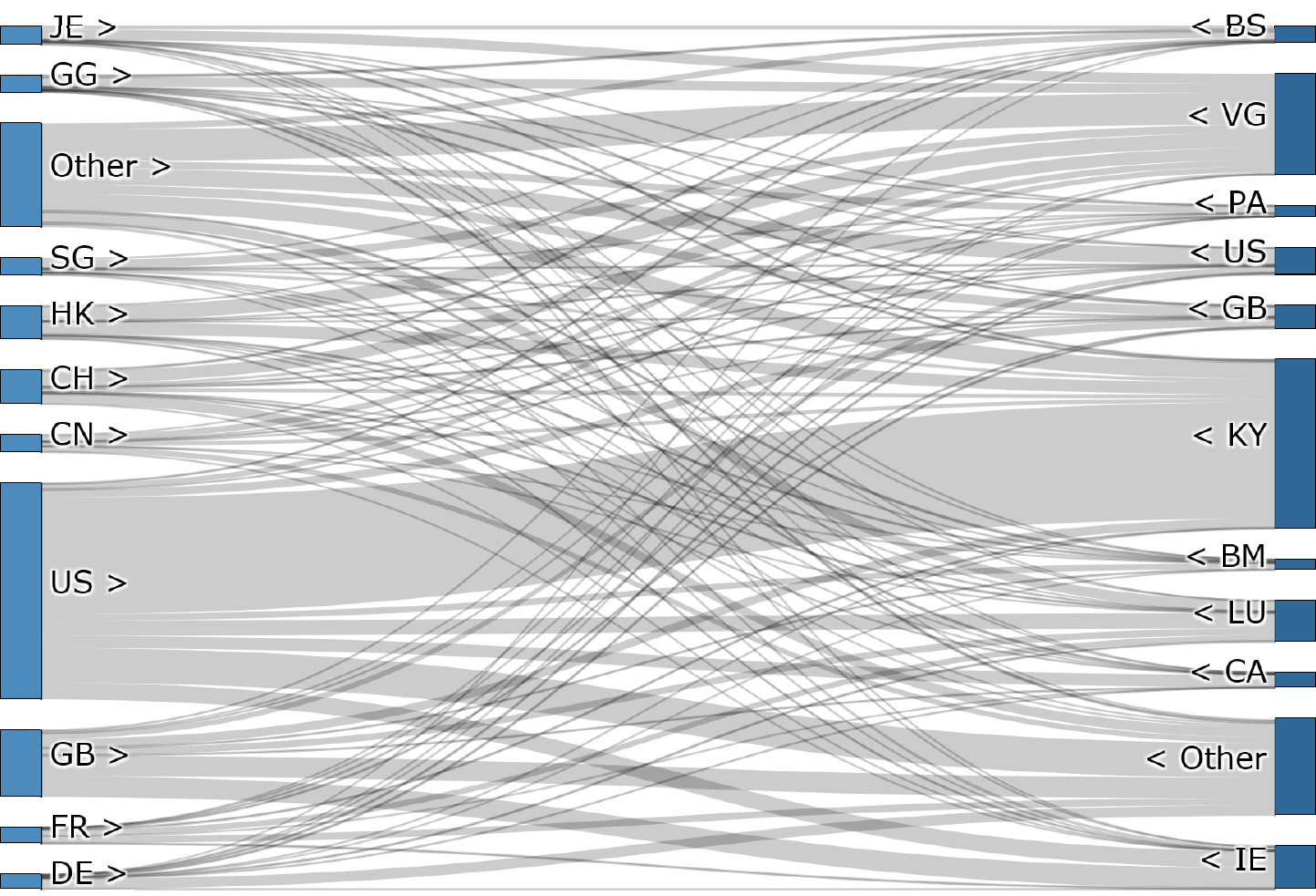}
    \caption{Most frequent headquarter (left) and legal address country (right) for companies where headquarter and legal address are in different countries.}
    \label{fig:headquarter_legal_flow}
\end{figure}

\begin{figure}[t]
    \centering
    \begin{subfigure}[t]{0.45\textwidth}
        \includegraphics[width=\textwidth]{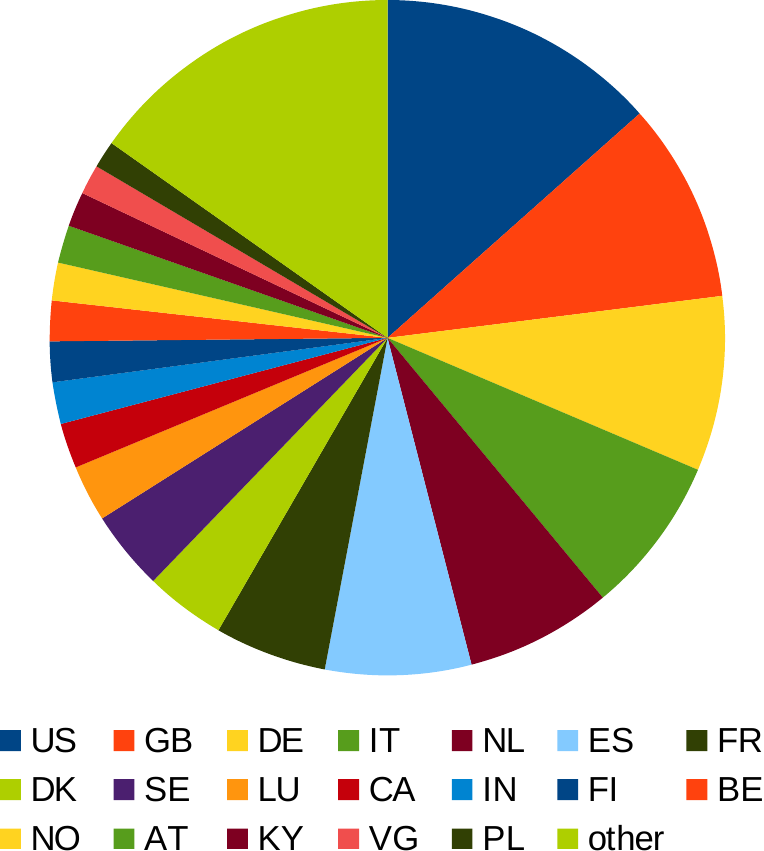}
        \caption{Legal address}
    \end{subfigure}
    $\quad\quad\quad$
    \begin{subfigure}[t]{0.45\textwidth}
        \includegraphics[width=\textwidth]{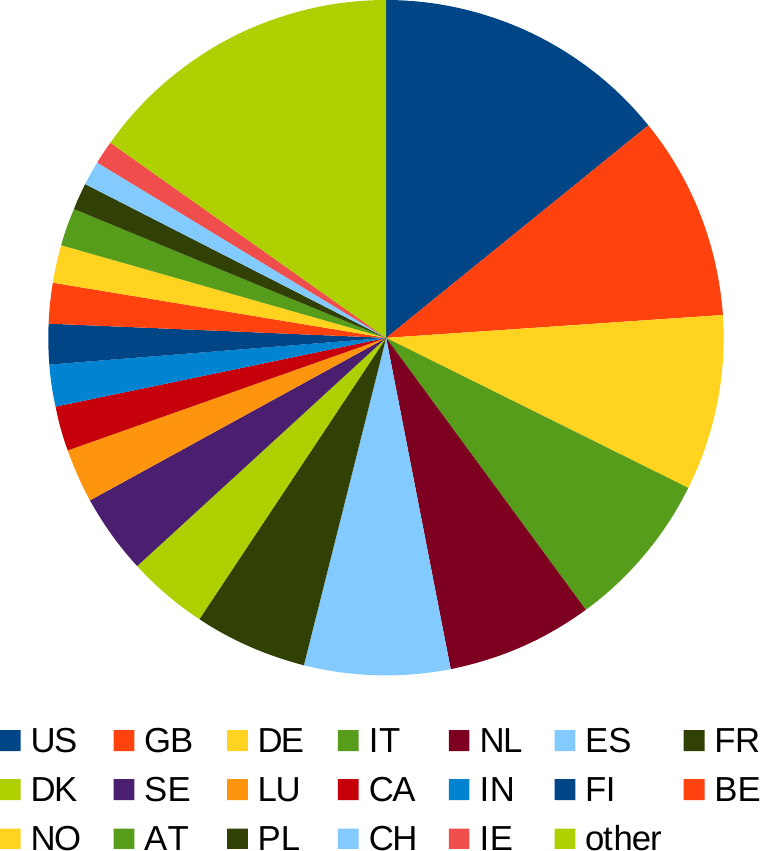}
        \caption{Headquarter address}
    \end{subfigure}
    \caption{Distribution of legal and headquarter addresses}
    \label{fig:addresses}
\end{figure}

Figure~\ref{fig:addresses} shows the distribution of legal and headquarter addresses. While the distribution among the top legal and headquarter addresses is similar, we can observe that two tax havens, i.e., Cayman Islands (KY) and British Virgin Islands (VG), appear among the top legal addresses, but not among the top headquarter addresses. For 36,400 of all companies in the graph (2.4\%), the headquarter and legal address country differ; the majority of legal addresses in this set are the Cayman Islands (9,838), British Virgin Islands (5,878), Ireland (2,496), and Luxembourg (2,389). The most common combination is a headquarter address in the USA and a legal address in the Cayman Islands, as depicted in Fig.~\ref{fig:headquarter_legal_flow}.

\begin{figure}[t]
    \centering
    \begin{subfigure}[c]{0.48\textwidth}
        \includegraphics[width=\textwidth]{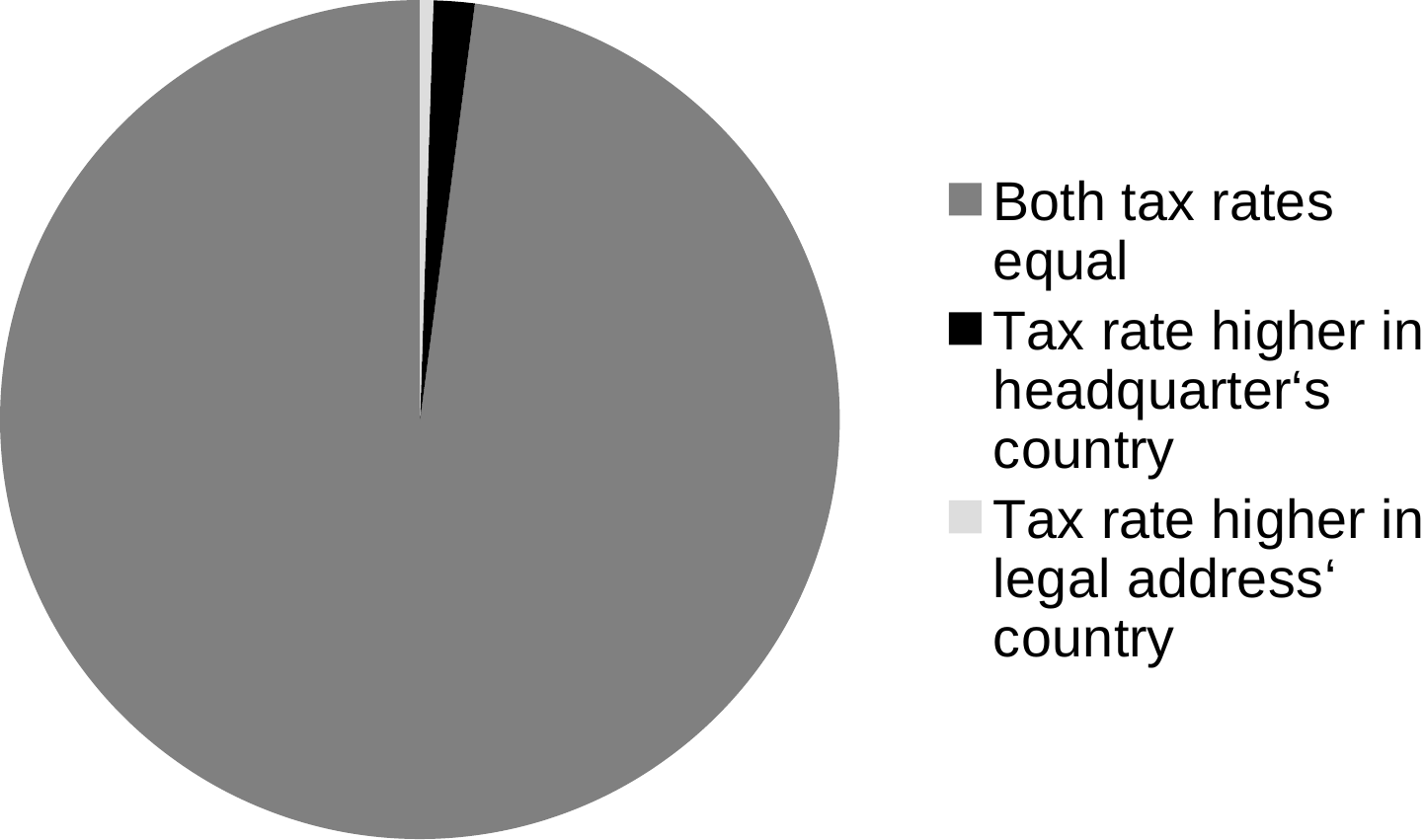}
    \end{subfigure}
    ~
    \begin{subfigure}[c]{0.48\textwidth}
        \includegraphics[width=\textwidth]{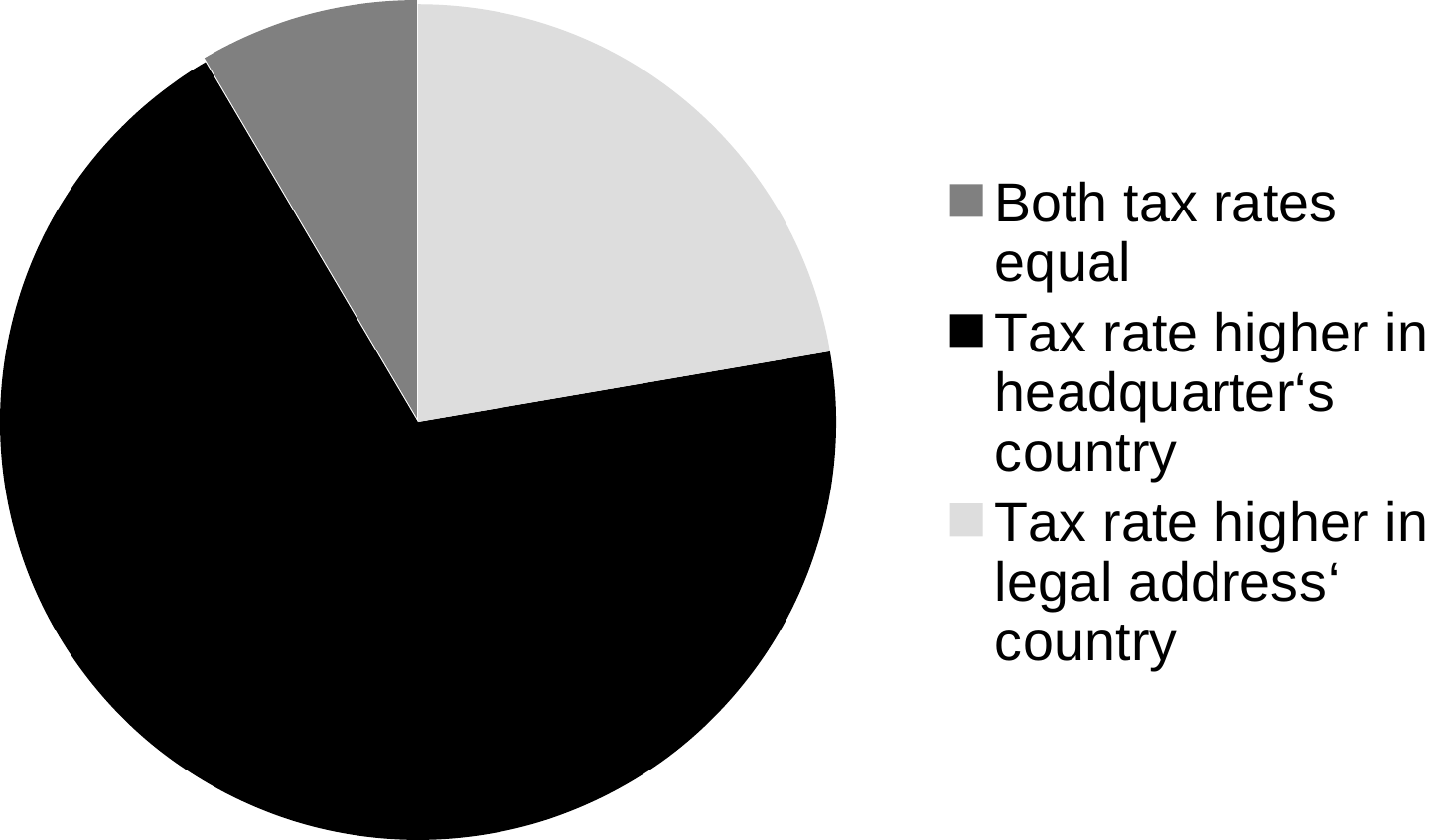}
    \end{subfigure}
    \\\vspace{0.25cm}
    \begin{subfigure}[c]{0.48\textwidth}
        \includegraphics[width=\textwidth]{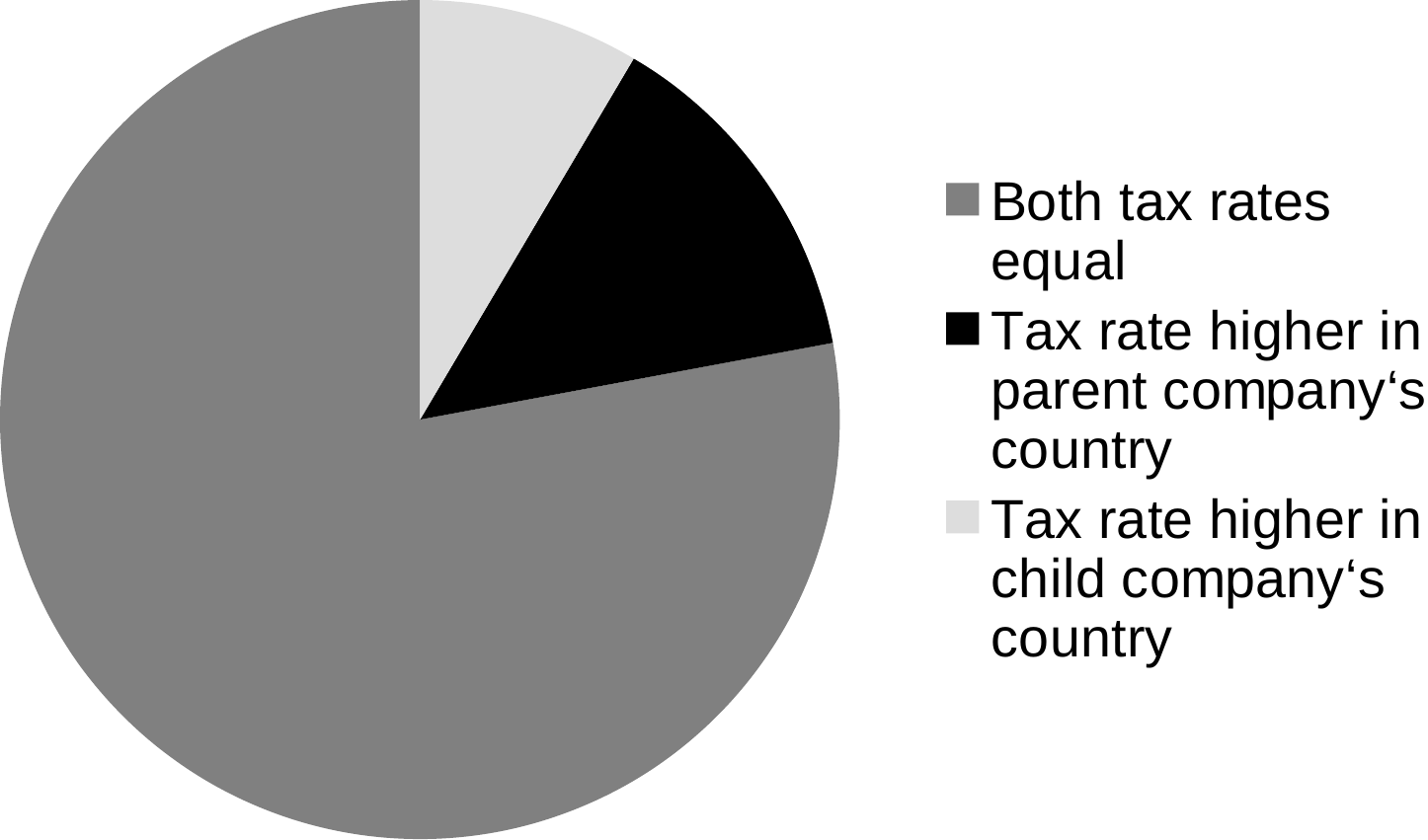}
    \end{subfigure}
    ~
    \begin{subfigure}[c]{0.48\textwidth}
        \includegraphics[width=\textwidth]{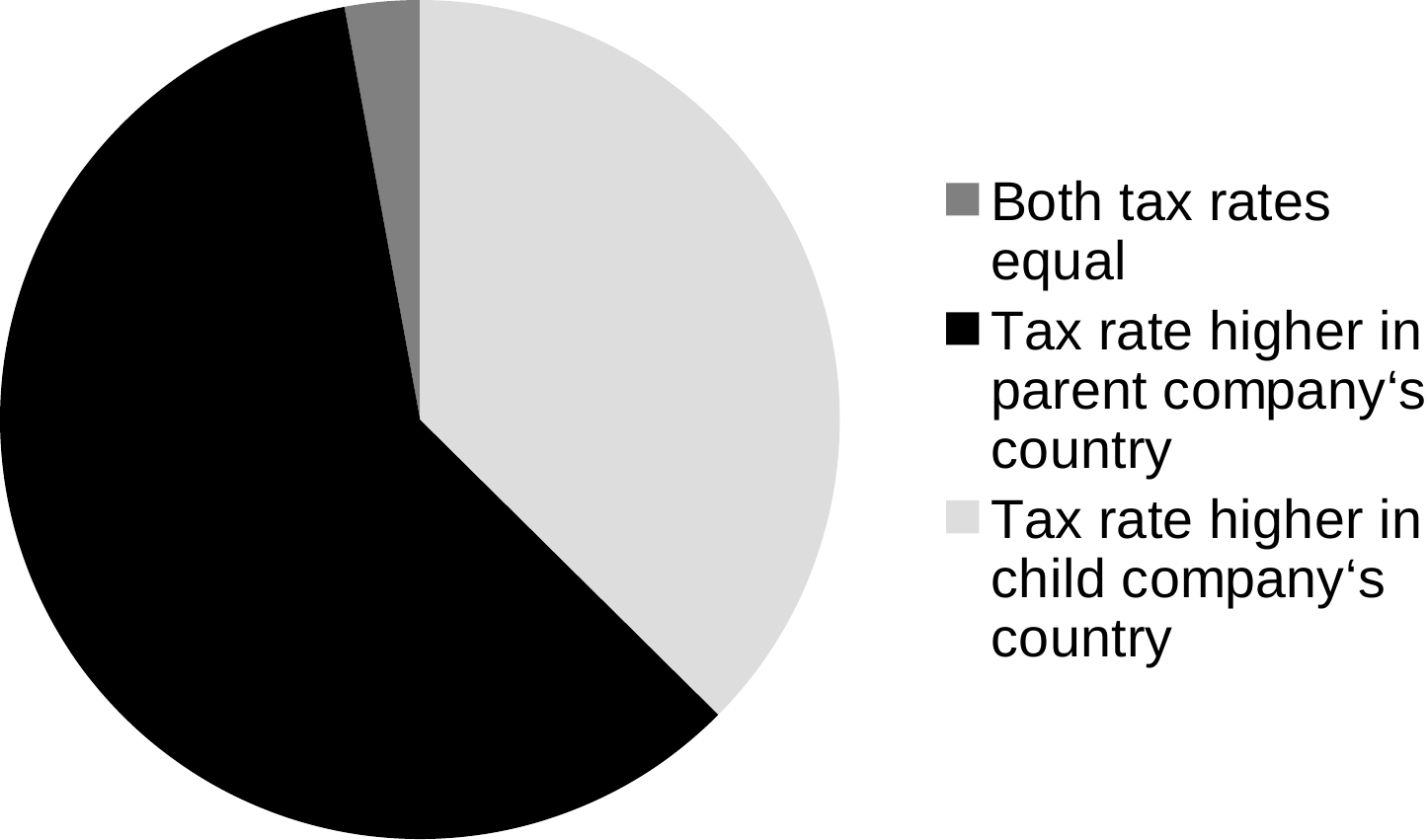}
    \end{subfigure}
    
    \caption{Corporate tax differences between headquarter and legal address (upper part) and parent and child companies' addresses (lower part). The left hand side diagrams depict all companies, the right hand side diagrams are filtered to those where the two countries are different.}
    \label{fig:tax_distribution}
\end{figure}

When comparing the corporate tax rates in the legal and headquarter addresses' countries, it can be observed that the corporate tax rate in the legal address country is, on average, 0.24 percentage points lower than in the headquarter's country. When considering only the 36,400 companies with differing addresses, that difference is even 10.5 percentage points. As depicted in Fig.~\ref{fig:tax_distribution}, companies having their headquarter and legal address in different countries have a higher tendency of using a legal address in a lower-tax country.

\begin{table}[t]
    \caption{Top 10 Companies with the most ultimate children}
    \label{tab:top10_ultimate_children}
    \centering
    \begin{tabular}{l|r}
         Company & No. of ultimate children\\
         \hline
         The Goldman Sachs Group, Inc. & 2,534\\
         Deutsche Bank Aktiengesellschaft & 885\\
         Morgan Stanley & 793\\
         Citigroup Inc. & 686\\
         Lloyds Banking Group PLC & 680\\
         Aegon N.V. & 629\\
         The Royal Bank of Scotland Group Public Limited Company & 496\\
         HSBC Holdings PLC & 472 \\
         Siemens Aktiengesellschaft & 455\\
         Societe Generale & 429\\
    \end{tabular}
\end{table}

\begin{figure}[t]
    \centering
    \includegraphics[width=\textwidth]{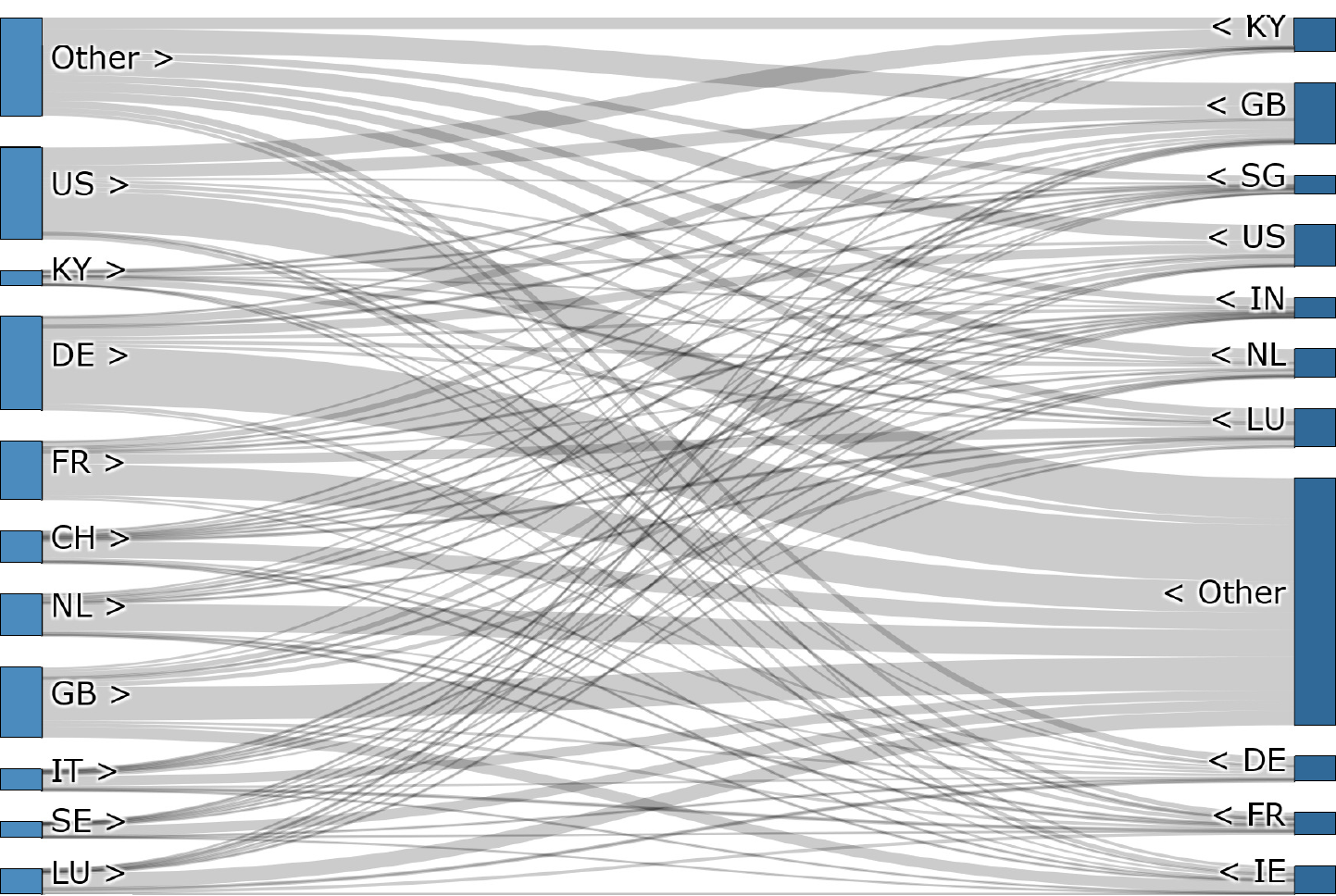}
    \caption{Most frequent headquarter of parent (left) and child company's legal address country (right) for multinational subsidiary relations.}
    \label{fig:subsidiary_flow}
\end{figure}

For subsidiary relations between companies, 35.7\% of those are multinational, i.e., the legal address country of the subsidiary and its affiliate differ. Fig.~\ref{fig:subsidiary_flow} depicts the most common relations for such multinational relationships. It can be observed that Ireland, India, and Singapore appear among the top 10 subsidiaries, but not among the top 10 parents.

When looking at the corporate tax rates for multinational companies, it can again be observed that the tax rate in which the subsidiary is located is typically lower than the one of the consolidating company. Across all subsidiary relations, the corporate tax rate in the child company's country is by 0.62 percentage points lower than in the parent company's country; if restricting this to multinational relations (i.e., where the parent and child company have their legal address in different countries), the difference is 2.46 percentage points.

\section{Usage Examples}
\label{sec:usage}
The knowledge graph can be used both for finding evidence for well-known tax avoidance strategies, as well as for searching for anomalies in the graph which hint at avoidance strategies not yet known. 

\subsection{Tax Avoidance Strategies}
Well-known strategies for tax avoidance can be observed in the graph and formulated as query patterns and graph queries.

\subsubsection{Double Irish with a Dutch Arrangement}
The Double Irish with a Dutch Arrangement uses in essence three companies: Two companies are located in Ireland (company A and C) and a conduit entity in the Netherlands (company B). Yet, the Irish fiscal authority considers only company A as taxable in Ireland, the second company is tax resident in a tax haven (company C). This allows to attribute all revenues to a tax haven (company C). \cite{fuest_profit_2013,kleinbard_stateless_2011} 

\begin{figure}[t]
    \centering
    \begin{subfigure}[b]{0.45\textwidth}
    \includegraphics[width=\textwidth]{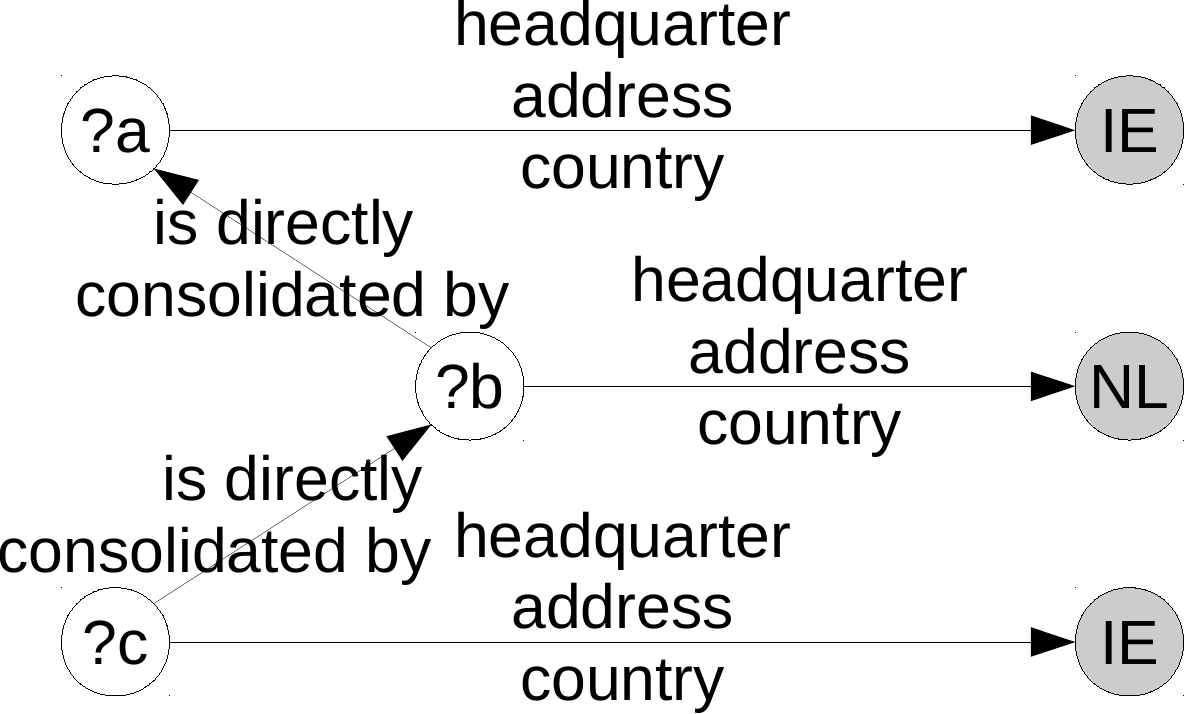}
    \end{subfigure}
    ~
    \begin{subfigure}[b]{0.5\textwidth}
    \begin{verbatim}SELECT *
WHERE {
  ?a tgp:isDirectlyConsolidatedBy+ ?b .
  ?b tgp:isDirectlyConsolidatedBy+ ?c .
  ?a tgp:headquartersAddressCountry ?tgc:IE .
  ?b tgp:headquartersAddressCountry ?tgc:NL .
  ?c tgp:headquartersAddressCountry ?tgc:IE .
}\end{verbatim}
    \end{subfigure}
    \caption{Double Irish Arrangement}
    \label{fig:double_irish}
\end{figure}

Fig.~\ref{fig:double_irish} depicts the query for a Double Irish with a Dutch Arrangement. Note that since further intermediate companies might be involved, we allow for chains of ownership by using \texttt{tgp:isDirectlyConsolidatedBy+}. 
Since the data in our knowledge graph is not complete, we could not find direct evidence for the Double Irish with a Dutch Arrangement construct. However, removing the last condition of the query (i.e., that company C has to have its headquarter in Ireland) yields 19 results (with the headquarter of C being located in countries such as the UK, the US, Japan, or Finland), which might hint at other variants of that tax planning strategy.

\subsubsection{Duck-Rabbit Construct}
Countries implement different legislative regulations which can have the unintended consequence that hybrid entities emerge. The OECD considers hybrid entities as firms with a dual residency and no country recognizes the entity as taxable. \cite{ludicke_tax_2014,oecd_neutralising_2015} These constructs are called \emph{duck-rabbit construct} in the following, named after the optical illusion in which some people see a duck, and some see a rabbit\footnote{\url{https://en.wikipedia.org/wiki/Rabbit-duck_illusion}}. The structure can be as follows: a company C in the Netherlands having the legal form of a BV (a private limited partnership) is the child of a company B in a tax haven, which in turn is the ultimate child of some international company A, usually located in the US. In that case, the Dutch laws consider B a company under US tax legislation, while the US laws consider B a company under Dutch tax legislation, which ultimately leads to the company being taxed in none of the two countries.

\begin{figure}[t]
    \centering
    \begin{subfigure}[b]{0.45\textwidth}
    \includegraphics[width=\textwidth]{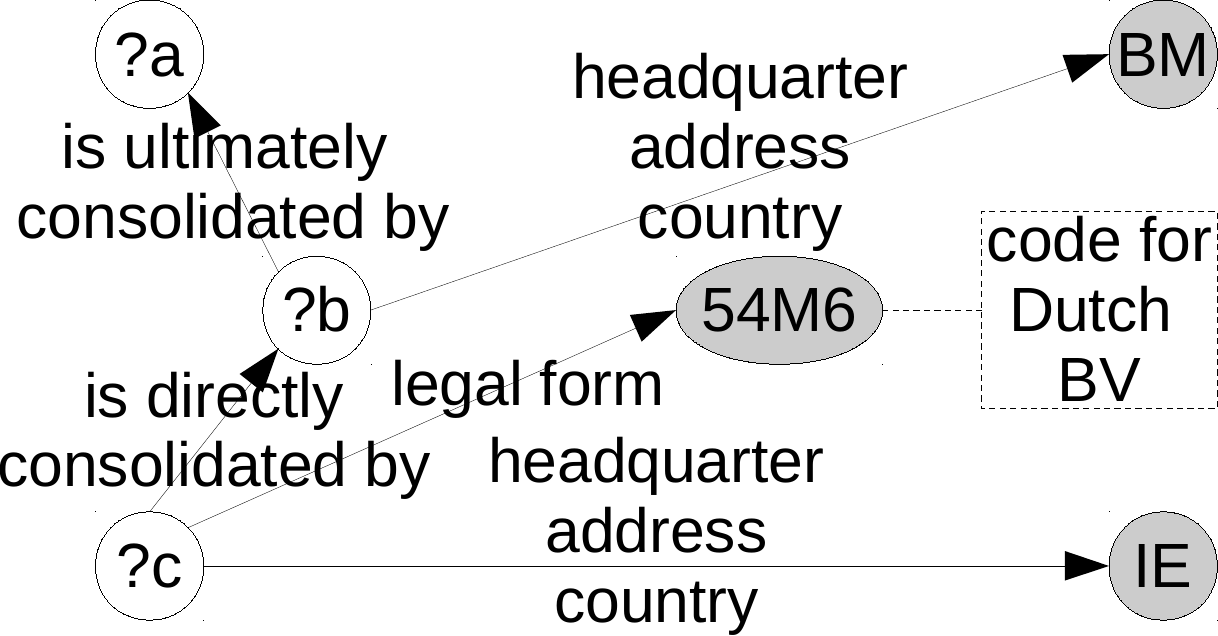}
    \end{subfigure}
    ~
    \begin{subfigure}[b]{0.5\textwidth}
    \begin{verbatim}SELECT *
WHERE {
    ?b tgp:headquartersAddressCountry tgc:BM .
    ?b tgp:isUltimatelyConsolidatedBy ?a .
    ?c tgp:headquartersAddressCountry tgc:NL .
    ?c tgp:isDirectlyConsolidatedBy ?b .
    ?c tgp:legalForm tglf:54M6 .
}     \end{verbatim}
    \end{subfigure}
    \caption{Duck Rabbit Construction}
    \label{fig:duck_rabbit}
\end{figure}

The corresponding graph pattern and query are shown in Fig~\ref{fig:duck_rabbit}. Running this query against the graph returns three constructs using the Bermudas and one using the Cayman Islands as an offshore tax haven. Among the former, there is also the game company \emph{Activision}, which has become one of the well-known examples for this kind of tax avoidance strategy.\footnote{\url{https://thecorrespondent.com/6942}}

\subsection{Graph Anomalies}
Since we included additional data about countries in our graph, we can use this as background information for further interesting observations \cite{ristoski2013analyzing}. One of those observations is the density of companies per state.

Table~\ref{tab:top10cpi} depicts the top 10 countries by companies per capita and companies per GDP. It can be observed that many known tax havens appear in the top positions, with some values being clearly out of range (e.g., Liechtenstein lists one company per three inhabitants).

In the table of companies per capita, Denmark appears to be a bit of an outlier at first glance. Digging a bit deeper, we found that private holding companies -- so called \emph{Anpartselskab} -- in Denmark are not taxed under certain conditions, and the creation of such companies is even advertised as a means for tax planning.\footnote{See, e.g., \url{https://www.offshorecompany.com/company/denmark-holding/}} While this finding was new to the domain experts in the team, and we have not been able to fully explain the Denmark anomaly, we can, as of today, only find that ``something is rotten in the state of Denmark.'' \cite{shakespeare1912hamlet}

\begin{table}[t]
    \caption{Top 10 countries by companies per inhabitants (top) and per GDP (in Million USD, bottom). Germany and USA are listed for comparison.}
    \label{tab:top10cpi}
    \centering
    \begin{tabular}{l|r|r}
Country	&	Population	&	Companies per capita \\
\hline
Liechtenstein	&	37,910	&	0.311 \\
Cayman Islands	&	64,174	&	0.237 \\
Luxembourg	&	607,728	&	0.063 \\
Isle of Man	&	84,077	&	0.036 \\
Bermuda	&	63,968	&	0.035 \\
Monaco	&	38,682	&	0.017 \\
Marshall Islands	&	58,413	&	0.017 \\
Seychelles	&	96,762	&	0.011 \\
Denmark	&	5,797,446	&	0.009 \\
Saint Kitts and Nevis	&	52,441	&	0.008 \\
\hdashline
\emph{Germany} & 82,927,922 & 0.002 \\
\emph{USA} & 327,167,434 & 0.001 \\
\hline\hline
Country & GDP & Companies per 1M GDP \\
\hline
Marshall Islands	&	221,278	&	4.59 \\
Cayman Islands	&	5,1413	&	2.96 \\
Liechtenstein	&	6,214	&	1.90 \\
Seychelles	&	1,590	&	0.66 \\
Belize	&	1,871	&	0.61 \\
Samoa	&	820	&	0.57 \\
Saint Vincent and the Grenadines	&	811	&	0.54 \\
Luxembourg	&	70,885	&	0.54 \\
Saint Kitts and Nevis	&	1,011	&	0.46 \\
Isle of Man	&	6,770	&	0.45 \\
\hdashline
\emph{Germany} & 3,947,620 & 0.03 \\
\emph{USA} & 20,544,343 & 0.01 \\
\end{tabular}
\end{table}

\begin{figure}[t]
    \centering
    \begin{subfigure}[t]{0.48\textwidth}
        \includegraphics[width=\textwidth]{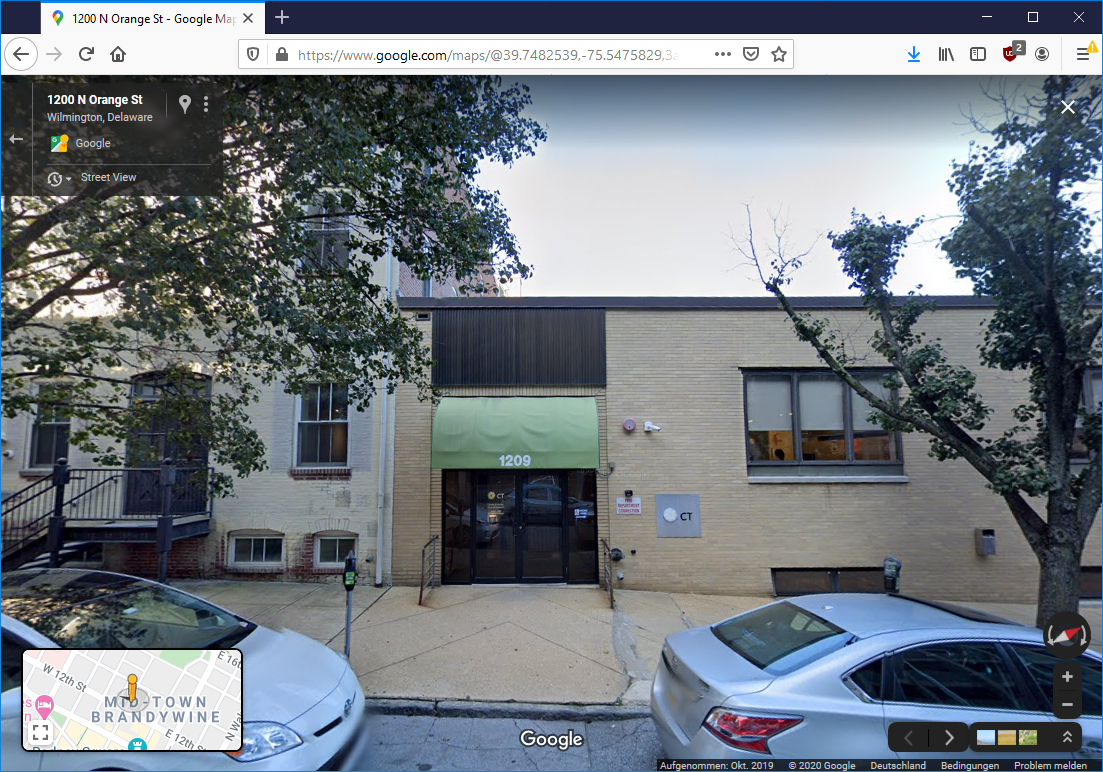}
        \caption{1209 Orange Street, Wilmington, Delaware, USA (14,551 companies)}
    \end{subfigure}
    ~
    \begin{subfigure}[t]{0.48\textwidth}
        \includegraphics[width=\textwidth]{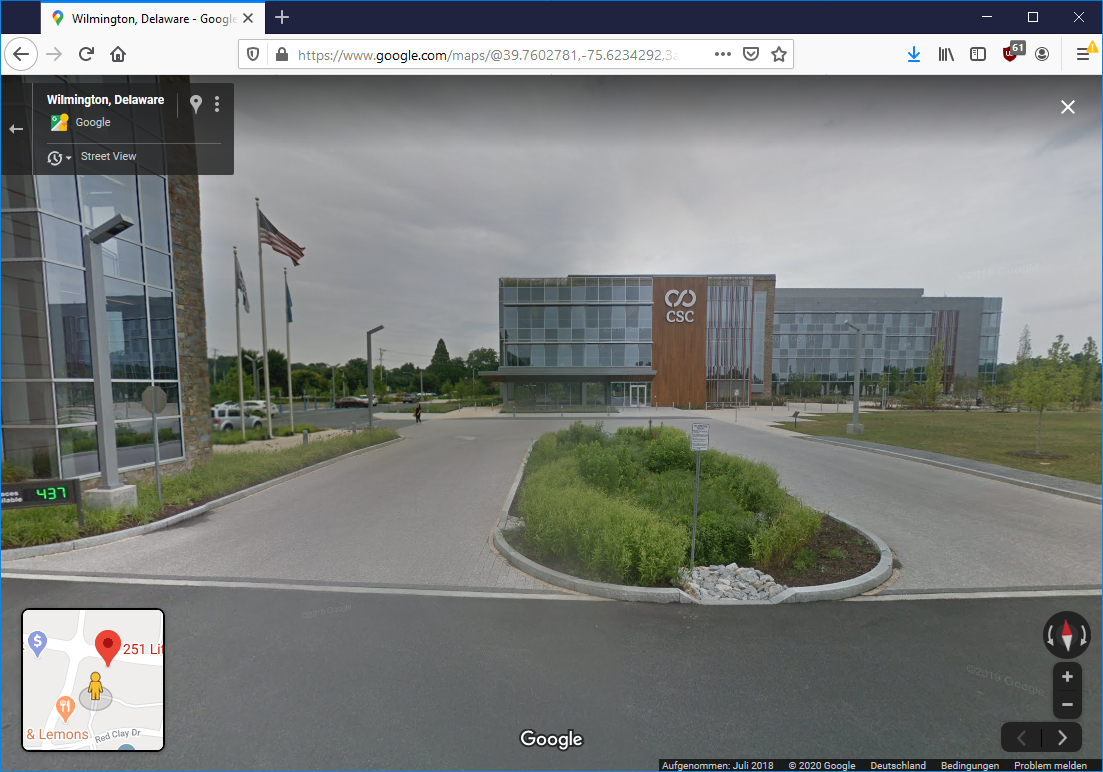}
        \caption{251 Little Falls Drive, Wilmington, Delaware, USA (11,207 companies)}
    \end{subfigure}
    \\\vspace{0.25cm}
    \begin{subfigure}[t]{0.48\textwidth}
        \includegraphics[width=\textwidth]{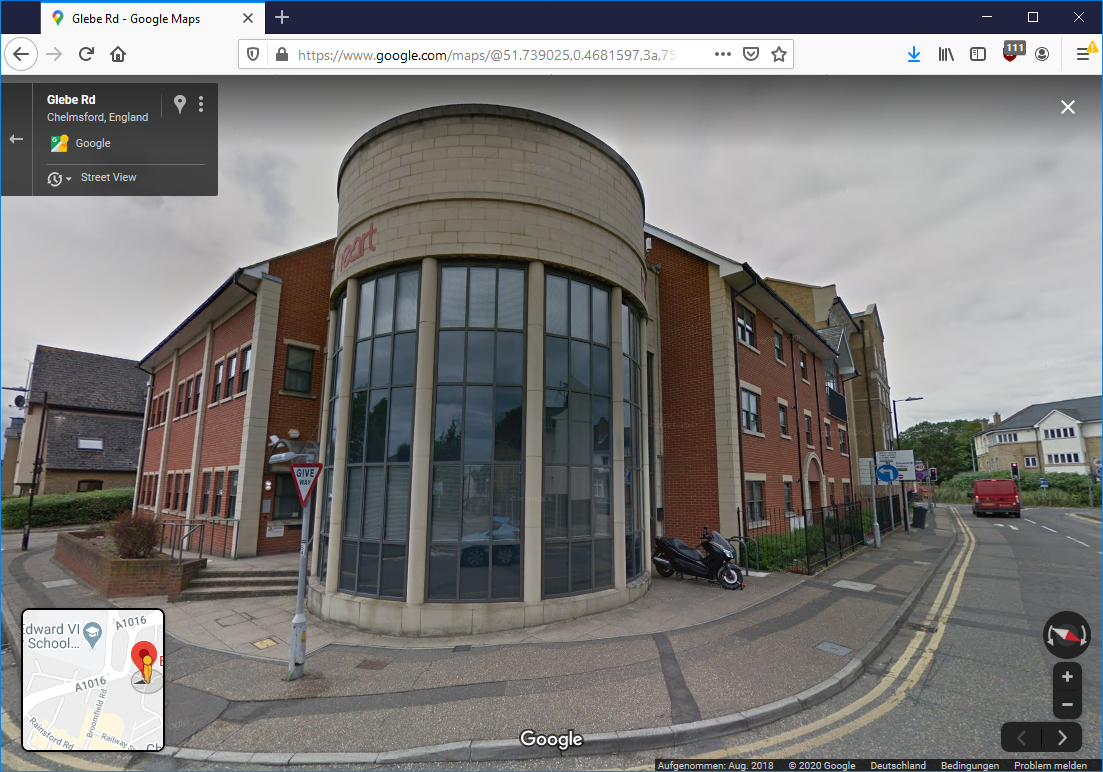}
        \caption{Eastwood House, Glebe Road, Chelmsford, CM1 1QW, Essex, United Kingdom (2,265 companies)}
    \end{subfigure}
    ~
    \begin{subfigure}[t]{0.48\textwidth}
        \includegraphics[width=\textwidth]{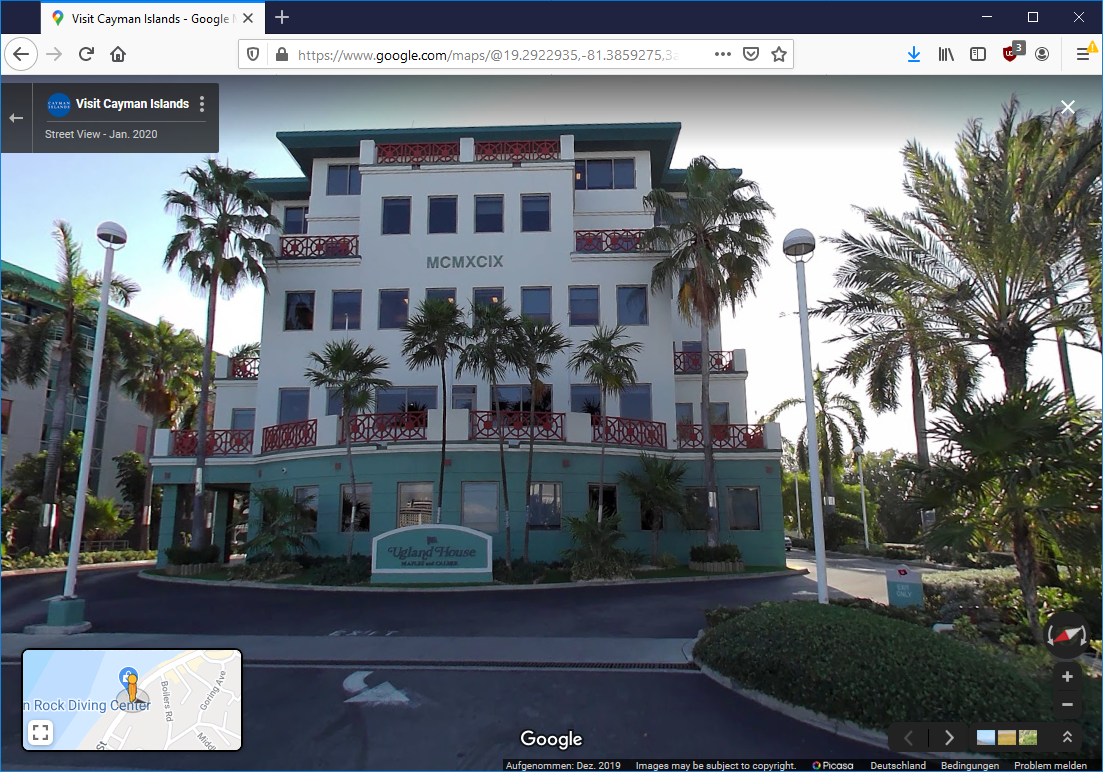}
        \caption{121 South Church Street, George Town, Cayman Islands (1,639 companies)}
    \end{subfigure}
\caption{Addresses with the highest frequency of being used as a legal address. Pictures from Google Street View.}
    \label{fig:addresses_examples}
\end{figure}

Another analysis we conducted is related to addresses with high concentrations of companies using that address as a legal address. There are quite a few addresses which are used as legal addresses by thousands of companies. Examples for such addresses are shown in Fig.~\ref{fig:addresses_examples}.

A particular observation of this analysis is that the two addresses most frequently used as legal addresses are in the state of Delaware, USA. We found that 36.7\% of all US companies in our knowledge graph have their legal address in Delaware, whereas the state only accounts for 0.29\% of the total US population. This phenomenon became known as the \emph{Delaware Loophole} \cite{wayne2012delaware} and is a result of the Delaware tax legislation, which does not charge income tax on companies not \emph{operating} in Delaware. \cite{dyreng_exploring_2013} Consequently, only 15.3\% of the companies having their legal address in Delaware also have their \emph{headquarter} in that state.

\subsection{Federated Querying}
Although, as discussed above, federated queries for combining data from our knowledge graph with data from Wikidata are not very fast and scalable, they are still possible. One example is to use the area of cities -- which is included in Wikidata but not in our KG -- and compute the density of companies by headquarter and legal address in each city. The rationale is that cities exposing an overly large density are suspicious, similar to the analysis of addresses above.

Figure~\ref{fig:example_federated} depicts an example for a federated query using Wikidata. The inner query collects all cities with a minimum number of companies using that city in their address, the outer query retrieves the area for those cities from Wikidata to compute the density of companies in those cities. Table~\ref{tab:top10density} shows the outcome of that query, showing the top 10 cities according to the density of headquarter and legal addresses registered. It can be observed that in both cases, Vaduz in Liechtenstein has the highest density of companies per square kilometer. For the density of legal addresses, Dover in Delaware shows up in the top list as another piece of evidence for the already mentioned \emph{Delaware Loophole}.\footnote{The top 10 lists, however, have to be taken with a grain of salt. For a city to appear in the top 10 list, it requires that (a) we are able to link it to Wikidata using the approach sketched in section~\ref{sec:kg}, and (b) it has to have its area as a value in Wikidata. Therefore, those lists cannot be considered complete.}

\begin{figure}[t]
    \centering
    \begin{verbatim}SELECT ?c ?count ?a (?count/?a as ?density) WHERE {
   { SELECT COUNT(?x) AS ?count ?c WHERE {
    ?x tgp:headquartersAddressCityID ?c .
    }
    GROUP BY ?c
    HAVING(COUNT(?x)>1000)
  } 
?c owl:sameAs ?wdc .
SERVICE <https://query.wikidata.org/bigdata/namespace/wdq/sparql> {
?wdc <http://www.wikidata.org/prop/direct/P2046> ?a}
} ORDER BY DESC(?density)\end{verbatim}
    \caption{Example for a federated query using Wikidata}
    \label{fig:example_federated}
\end{figure}

\begin{table}[t]
    \caption{Cities with largest densities of companies having registered their headquarter (upper half) and legal address (lower half)}
    \label{tab:top10density}
    \centering
    \begin{tabular}{l|l|r|r|r}
         City & Country & No. of companies & Area in sq. km. & Density \\
         \hline
        Vaduz	&	Liechtenstein	&	9021	&	17.30	&	521.45\\
        Puteaux	&	France	&	1334	&	3.19	&	418.18\\
        Paris	&	France	&	16276	&	105.40	&	154.42\\
        Geneva	&	Switzerland	&	2153	&	15.92	&	135.24\\
        Brussels	&	Belgium	&	3756	&	33.00	&	113.82\\
        Copenhagen	&	Denmark	&	7356	&	86.70	&	84.84\\
        Barcelona	&	Spain	&	8305	&	101.30	&	81.98\\
        Milan	&	Italy	&	12563	&	181.67	&	69.15\\
        Zug	&	Switzerland	&	1027	&	21.61	&	47.52\\
        Nicosia	&	Cyprus	&	2276	&	51.06	&	44.58\\
         \hline
		Vaduz	&	Liechtenstein	&	8460	&	17.30	&	489.02\\
		Puteaux	&	France	&	1218	&	3.19	&	381.82\\
		Dover	&	USA (Delaware)	&	11268	&	60.82	&	185.28\\
		Paris	&	France	&	16253	&	105.40	&	154.20\\
		Brussels	&	Belgium	&	4213	&	33.00	&	127.67\\
		Geneva	&	Switzerland	&	1522	&	15.92	&	95.60\\
		Copenhagen	&	Denmark	&	7432	&	86.70	&	85.72\\
		Barcelona	&	Spain	&	8095	&	101.30	&	79.91\\
		Milan	&	Italy	&	12599	&	181.67	&	69.35\\
		Zug	&	Switzerland	&	1052	&	21.61	&	48.68\\
    \end{tabular}
\end{table}

\section{Related Work}
\label{sec:related}
Parts of GLEIF, which we also used in this paper, have already been ported to an RDF representation and made available as a Linked Data endpoint \cite{trypuz2016general}. However, the most important information for our use case -- i.e., parent and child relations between companies -- are not included in that representation.

Other approaches are restricted to single branches and/or countries, and thus would not allow for an analysis like the one conducted in this paper. An example for a branch specific solution is discussed in~\cite{fan2018ontology}, where the authors build a populated ontology of bank holding companies and their ownership relations is introduced. The authors build an ontology and populated it from the Federal Reserve's public National Information Center (NIC) database\footnote{\url{https://www.ffiec.gov/NPW}}. Examples for country-specific solutions include a knowledge graph of Chinese companies~\cite{ma2015knowledge}, and a Linked Data endpoint of French business register data~\cite{el2019modeling}. The euBusinessGraph \cite{roman2020enabling} project publishes data about businesses in the EU, but does not contain relationships between companies. Those datasets are often very detailed, but are of limited use for analyzing the taxation of multinational companies.

\begin{figure}[t]
    \centering
    \includegraphics[width=\textwidth]{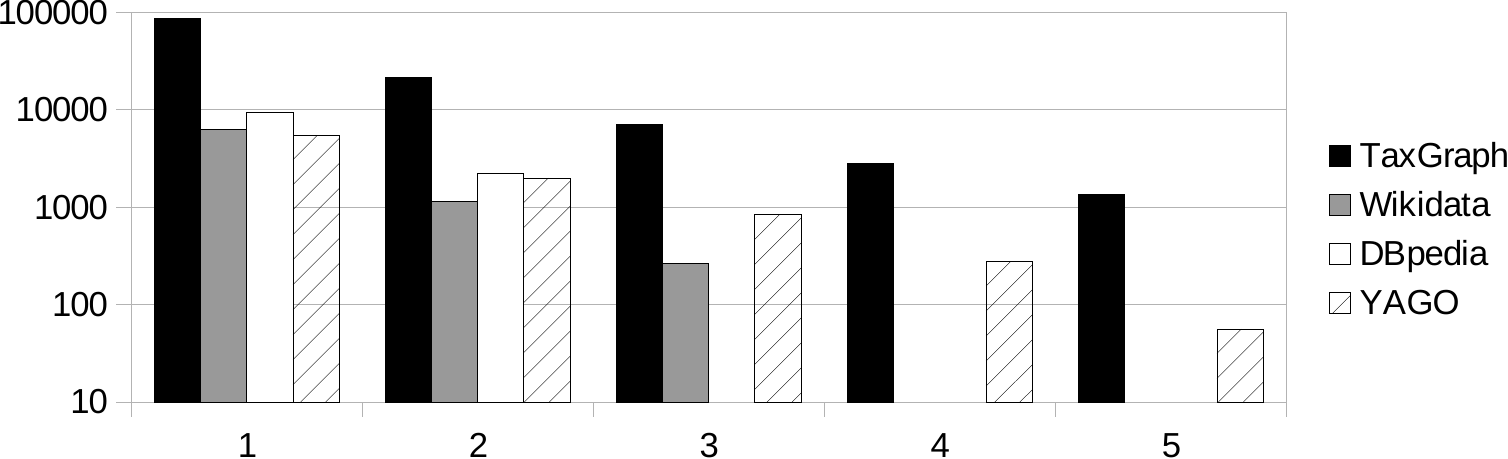}
    \caption{Chains of subsidiaries by number of hops}
    \label{fig:comparison_kgs}
\end{figure}

In addition to specific datasets, many cross-domain knowledge graphs also contain information about companies \cite{ringler2017one}. Hence, we also looked at such knowledge graphs as potential sources for the analysis at hand. However, since we need information not only for the main business entities, but also for smaller subsidiaries in order to identify tax compliance issues, we found that the information contained in those knowledge graphs is not sufficient for the task at hand. In Wikidata \cite{vrandevcic2014wikidata}, DBpedia \cite{lehmann2015dbpedia}, and YAGO \cite{mahdisoltani2013yago3}, the information about subsidiaries is at least one order of magnitude less frequent than in the graph discussed in this paper, as shown in Fig.~\ref{fig:comparison_kgs}: Especially  longer chains of subsidiary relations, which are needed in our approach, are hardly contained in public cross-domain knowledge graphs.

In the tax accounting literature several scholars have already used data on multinational corporations to analyze the behavior of firms. It has been shown that some firms fail to publicly disclose subsidiaries that are located in tax havens~\cite{dyreng_strategic_2018}. In~\cite{de_simone_real_2019}, the authors used very detailed data on the structure of multinational corporations to show that the introduction of public country by country reporting -- the requirement to provide accounting information for each country a firm operates in to tax authorities -- leads to a reduction in tax haven engagement.

A different strand of the tax literature has looked at networks of double tax treaties. Double tax treaties are in general bilateral agreements between countries that lower cross-border taxes in case of international transactions of multinational corporations. This literature on networks shows that some countries are strategically good choices for conduit entities to relocate profits and minimize cross-border taxation \cite{hong_tax_2016,vant_riet_profitable_2015}.

\section{Conclusion and Outlook}
\label{sec:conclusion}
In this paper, we have introduced a knowledge graph for multinational companies and their interrelations. We have shown that the graph allows for finding companies using specific constructs, such as well-known aggressive tax planning strategies, as well as for identifying further anomalies.

Our current knowledge graph uses company data from GLEIF, which is openly available and encompasses about 1.5M business entities. There are other (non-open) databases such as ORBIS~\cite{ribeiro2010oecd}, which contain even more than 40M business entities, but their licenses do not allow for making them available as a public knowledge graph. For the future, we envision the dual development of an open and a closed version of the graph, the latter based on larger, but non-public data. On the larger graph, we expect to find more evidence for known tax planning strategies and a larger number of interesting anomalies.

Another interesting source of information would be the mining of up to date information from news sites such as Reuters or Financial Times. This would allow feeding and updating the KG with recent information, and to directly rate events in the restructuring of multinational companies in the light of whether or not it is likely that those events happen for reasons of tax planning.

Apart from increasing the mere size of the graph, we also plan to include more diverse data in the graph. For example, adding branch information for companies would allow for more fine-grained analyses finding tax planning strategies, which are specific to particular branches. Further data about companies could include the size of companies (in terms of employees), or other quantitative revenue data mined from financial statements, and a detailed hierarchy of subsidiary relations describing the relations more closely (e.g., franchise, licensee, holding). 

A particular challenge lies in the more detailed representation of taxation legislation. For the moment, we have only included average corporate tax rates as a first approximation, but having more fine grained representations in the knowledge graph would be a clear improvement. However, this requires some up-front design considerations, since the ontological representation of tax legislation is not straight forward.

\bibliographystyle{splncs04}
\bibliography{bibliography}

\end{document}